\documentclass{llncs}
\usepackage{makeidx}  % allows for indexgeneration
\usepackage{comment}
\usepackage{xcolor}
\usepackage{graphicx}
\usepackage{booktabs}
\usepackage{gb4e}
\usepackage{amsmath}
\usepackage{suffix}
\usepackage{xspace}
\usepackage{framed}

\newcommand{\phil}[1]{\commment{Philipp}{#1}}
\newcommand{\hen}[1]{\commment{Hendrik}{#1}}
\newcommand{\rom}[1]{\commment{Roman}{#1}}
\newcommand{\mat}[1]{\commment{Matthias}{#1}}

\WithSuffix\newcommand\phil*[1]{\annotategreen{Philipp}{#1}}
\WithSuffix\newcommand\rom*[1]{\annotateblue{Roman}{#1}}
\WithSuffix\newcommand\mat*[1]{\annotatered{Matthias}{#1}}
\WithSuffix\newcommand\hen*[1]{\annotateyellow{Hendrik}{#1}}
\begin{document}
\frontmatter          % for the preliminaries
\pagestyle{headings}  % switches on printing of running heads

\mainmatter              % start of the contributions
\title{Predicting Disease-Gene Associations using Cross-Document
  Graph-based Features}
\titlerunning{Predicting Disease-Gene Associations}  % abbreviated title (for running head)
\author{Hendrik {ter Horst}\inst{1} \and Matthias Hartung\inst{1} \and Roman Klinger\inst{1,2} \and\\
Matthias Zwick\inst{3} \and Philipp Cimiano\inst{1}}
\authorrunning{Hendrik ter Horst et al.} % abbreviated author list (for running head)
\tocauthor{Hendrik ter Horst, Matthias Hartung, Roman Klinger, Matthias Zwick and Philipp Cimiano}
\institute{Semantic Computing Group, CITEC, Bielefeld University\\
\email{\{hterhors,mhartung,cimiano\}@techfak.uni-bielefeld.de}
\and
Institute for Natural Language Processing, University of Stuttgart\\
\email{roman.klinger@ims.uni-stuttgart.de}\\
\and
Research Networking, Boehringer Ingelheim Pharma GmbH \& Co. KG\\
\email{matthias.zwick@boehringer-ingelheim.com}}

\maketitle

\begin{abstract}
In the context of personalized medicine, text mining methods pose an interesting option for identifying disease-gene associations, as they can be used to generate novel links between diseases and genes which may complement knowledge from structured databases. The most straightforward approach to extract such links from text is to rely on a simple assumption postulating an association between all genes and diseases that co-occur within the same document. However, this approach (i) tends to yield a number of spurious associations, (ii) does not capture different relevant types of associations, and (iii) is incapable of aggregating knowledge that is spread across documents. Thus, we propose an approach in which disease-gene co-occurrences and gene-gene interactions are represented in an RDF graph. A machine learning-based classifier is trained that incorporates features extracted from the graph to separate disease-gene pairs into valid disease-gene associations and spurious ones. On the manually curated \textit{Genetic Testing Registry}, our approach yields a 30 points increase in $\text{F}_1$ score over a plain co-occurrence baseline.
  % We present a system and a machine learning model for predicting disease-gene
  % associations from biomedical text, using a combination of features
  % based on gene-disease co-occurrences and gene interactions that are
  % represented in a graph database. The latter group of features is
  % designed as to aggregate gene interaction knowledge across
  % individual documents. Our experimental results show that both types
  % of features are effective, as they contribute to outperform a plain
  % co-occurrence baseline by more than 30 points in $F_1$ score when
  % being evaluated against a manually curated database of gene-disease
  % associations.\rom{This abstract needs to be rewritten: Extracting
  %   gene/protein-disease relations is necessary as structured
  %   databases are potentially incomplete. Single document-based
  %   inference might not be sufficient (prominent example: Swanson 1986:
  %   10.1353/pbm.1986.0087). How to combine knowledge is difficult,
  %   there graph-based features. Short statement about result.}

\end{abstract}
\section{Introduction}

%personalized medicine  is the use of patient- specific information and  biomarkers  to make more informed  choices regarding the optimal therapeutic treatment regimen for that  patient

Most current approaches in personalized medicine, irrespective of
particular treatment modalities (e.g., small molecules, biologics,
novel approaches like gene therapy), are centered around modulating a
gene in order to modulate aspects of a disease
\cite{Kisoretal2014}. Therefore, the detection of disease-gene links
is an important starting point in drug discovery.

%Gene-disease links are also interesting as hints towards potential adverse effects.

Text mining methods pose an interesting option for identifying
disease-gene links, as they can be used to generate new target (and
therefore treatment) hypotheses and, in combination with experimental data, support the prioritization of research aimed at the discovery of new drug targets. Until now, 
text mining methods for disease-gene associations mostly rely on the degree of textual co-occurrence \cite{quan2014gene}. While those approaches are largely reliable in detecting well-known links for well-known
diseases \cite{al2005new,chun2006extraction,pletscher2014diseases}, such well-known links are of minor interest in drug research, as they do not support the discovery of new targets. Such novel targets are
difficult to detect, as they often require the aggregation of evidence across individual documents; at the same time,
they potentially shed light on yet unknown disease-gene links. % that may trigger innovative developments.

In this paper, we propose a classification model for predicting novel 
disease-gene associations from biomedical text. The model combines
\textit{private} (intra-document) as well as \textit{public} (cross-document) knowledge as defined by Swanson et al.\
\cite{swanson1996undiscovered} in terms of features based on local
co-occurrences within documents and relations between diseases and genes that have been aggregated across individual documents. In an evaluation against an existing database, we address the following research questions: (i) Can such a combined
model outperform a purely co-occurrence-based approach? (ii) What is
the impact of features measuring the connectivity of diseases and
genes? (iii) What is the impact of graph-based features capturing
interactions between genes across documents?

\section{System Architecture}\label{sec:sysarc}
\begin{figure}[th]
 \centering
 \includegraphics[width=0.9\textwidth]{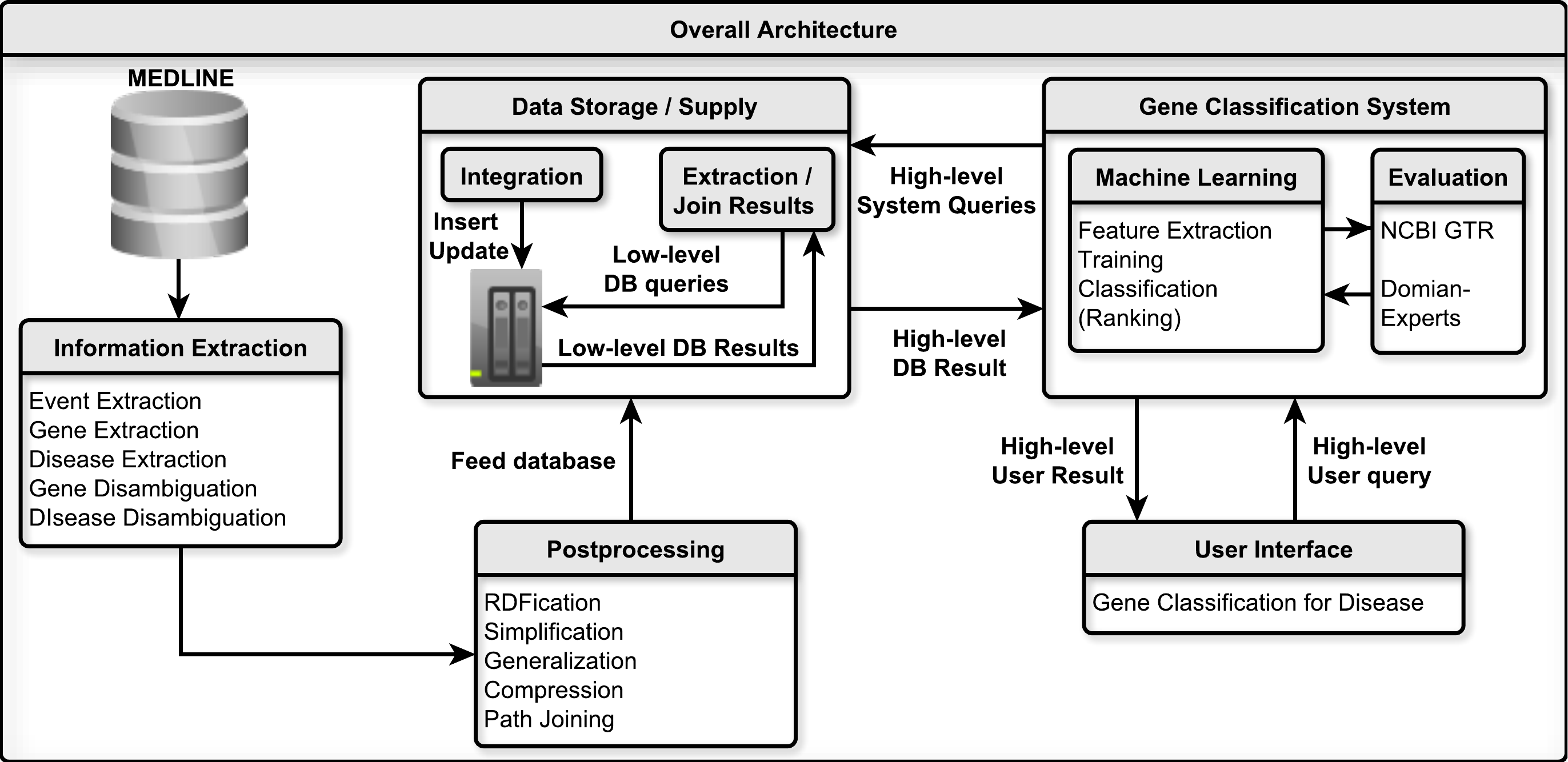}
 \caption{Overview of system architecture}
 \label{fig:architecture}
\end{figure}

Our system architecture consists of the following components (cf.\ Fig.\ \ref{fig:architecture}):

\begin{enumerate}
\item \textbf{Medline Corpus}: All analyses are done on Medline\footnote{\url{http://www.ncbi.nlm.nih.gov/pubmed}}, comprising a total of 21.5M abstracts.
\item \textbf{Information Extraction:} We rely on existing information extraction systems to identify disease names, genes/proteins and interactions between them. 
For \textbf{disease recognition}, we use a state-of-the-art CRF tagger \cite{Klingeretal2007} that has been trained on the NCBI Disease Corpus \cite{DoganLu2012}. Using the normalization procedure described in \cite{terHorst2015}, disease names are reduced to a vocabulary of approx.\ 15K unique identifiers extracted from MeSH\footnote{\url{https://www.nlm.nih.gov/mesh/}} and OMIM\footnote{\url{http://omim.org}}. 
For the \textbf{identification of genes/proteins and their
  interactions}, we rely on TEES \cite{bjorne2009extracting}, a
state-of-the-art event extraction system that has been tailored to the
detection of molecular interactions from biomedical
text. All extracted genes are normalized using GeNo\ \cite{geno} and afterwards filtered by human genes using taxonomic information from EntrezGene \cite{maglott2011entrez}.
\begin{figure}[th]
\centering
\includegraphics[width=\textwidth]{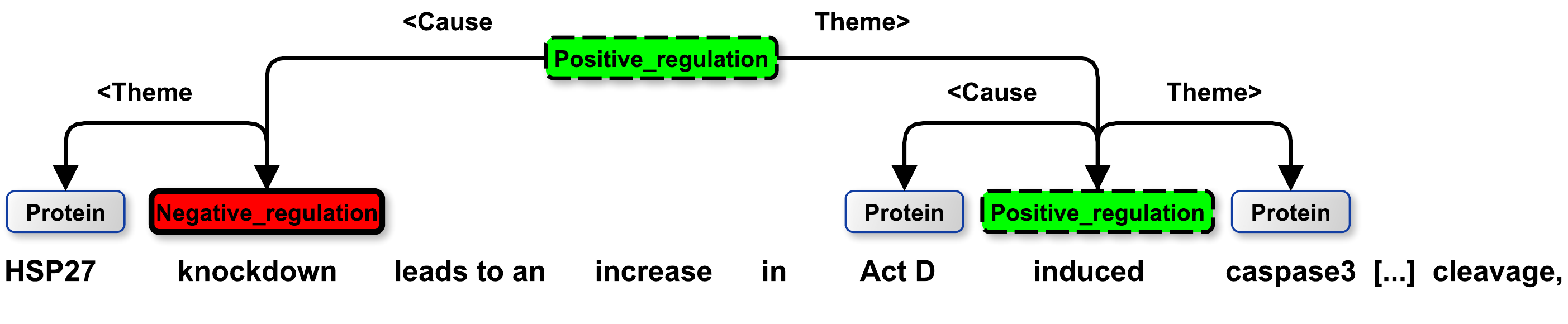}
\caption{TEES representation of an example sentence from \cite{Maetal2013}.}
\label{fig:tees-example}
\end{figure}
\item \textbf{Postprocessing:} The complexity of the interaction
  graphs produced by TEES (see Fig.\ \ref{fig:tees-example}) is
  incrementally reduced. Fig.\ \ref{fig:pp-example} gives a running
  example of the individual steps described in the
  following:
\begin{enumerate}%[topsep=0pt,itemsep=-1ex,partopsep=1ex,parsep=1ex]
\item \textbf{RDFication:} TEES graphs are broken down into binary relations by selecting all shortest paths that connect two proteins with respect to their semantic relation. These are represented as one RDF triple connecting two proteins. The path between the two proteins is serialized as a string and used as the name of an RDF property connecting both proteins.
\item \textbf{Simplification:} Semantic role information (i.e., \textsc{cause} and \textsc{theme}) is omitted. The direction of interactions is still captured in the directed edges of the graph.%, while causality is not of primary interest in this work.
\item \textbf{Generalization:} The TEES extraction scheme, originally consisting of 9 relations, was reduced to five relations after discussion with a domain expert: \textit{Expression}, \textit{Catabolism}, \textit{Localization}, \textit{Binding} and \textit{Regulation}.
\item \textbf{Compression:} Consecutive occurrences of identical relations within a path signature are compressed by reducing them to one relation.
\item \textbf{Path joining:} To extract longer dependencies between genes/proteins as well, we join paths connecting two genes up to a distance of two edges.
%However, the system is not limited in the number of edges.
 The join of the paths is serialized again, the above post-processing steps are applied and the results are stored as RDF triples. We refer to such serialized paths as \emph{path signatures}. 
\label{ref:pathsigdef}
\end{enumerate}

\begin{figure}
\centering
\footnotesize
\begin{framed}
\begin{itemize}
\item[(a)]HSP27 -- \textsc{theme}:Neg\_reg:\textsc{cause}:Pos\_reg:\textsc{theme}:Pos\_reg:\textsc{cause} -- ActD \\
HSP27 -- \textsc{theme}:Neg\_reg:\textsc{cause}:Pos\_reg:\textsc{theme}:Pos\_reg:\textsc{theme} -- caspase3\\
ActD -- \textsc{cause}:Pos\_reg:\textsc{theme} -- caspase3\\
\item[(b)]HSP27 -- Neg\_reg:Pos\_reg:Pos\_reg -- ActD \\
HSP27 -- Neg\_reg:Pos\_reg:Pos\_reg -- caspase3\\
ActD -- Pos\_reg -- caspase3
\item[(c)]HSP27 -- Reg:Reg:Reg -- ActD \\
HSP27 -- Reg:Reg:Reg -- caspase3\\
ActD -- Reg -- caspase3
\item[(d)]HSP27 -- Reg3 -- ActD \\
HSP27 -- Reg3 -- caspase3\\
ActD -- Reg -- caspase3
\end{itemize}
\end{framed}
\caption{Example of the post-processing procedure applied to TEES
  events. Paths in (a) are equivalent to the original TEES output
  (cf.\ Fig.\ \ref{fig:tees-example}), (d) shows the result after
  post-processing, without path joining being applied.}
\label{fig:pp-example}
\end{figure}

\item \textbf{Database}: The results of all the steps described above are stored in an RDF database, \textit{Blazegraph}\footnote{\url{http://www.blazegraph.com}}.
\item \textbf{Gene Classification System:} Given a disease as input, protein candidates are classified as to whether or not they interact with that disease. The classifier relies on features extracted for each pair of disease and protein. Being our main contribution, this component is described in the next section. 
%\item \textbf{User Interface:} The frontend by which the user can interact with the system.\hen{Could be deleted to save space? Maybe adjust System-architecture image.}
\end{enumerate}

\section{Gene Classification System}

Our gene classification system takes a disease as input and predicts,
for each gene in the database, whether it interacts with the given
disease or not. The classifier is implemented as a Support Vector
Machine relying on features that are extracted for each disease-gene
pair. Seven feature groups are defined in total, which can be divided
into \textit{co-occurrence-based (CBF)} and \textit{graph-based
  (GBF)} features, as described below.

Our notation is as follows: Let $D$ be the set of
all diseases and $G$ the set of all genes in the database\footnote{$D$ and $G$ comprise all diseases and genes recognized during preprocessing the Medline corpus (cf.\ Section \ref{sec:sysarc}). This amounts to 7.640 diseases and 11.201 genes, in total.} and $P$ a vocabulary of predicates denoting semantic relations between them. Then, $T$ denotes the set of all triples in the database, such that:
$T \subset (D \times P \times G) = \{\langle d,p,g\rangle | p=\mathit{coocc}\} \cup \{\langle g,p,g'\rangle | p=\mathit{interact}\}$. 

\subsection{Co-occurrence-based Features}\label{sec:cooccurrencebased}
CBF features are based on the co-occurrence between diseases and
genes. We consider $T_d \subset T = \{\langle d', \mathit{coocc}, g\rangle | d'=d\}$, the set of all genes co-occurring with a particular disease $d$, and analogously $T_g \subset T = \{\langle d, \mathit{coocc}, g'\rangle | g'=g\}$. Moreover, $T_{dg} \subset T= \{ \langle d', \mathit{coocc}, g'\rangle | d'=d, g'=g\}$ denotes the set of all co-occurrences of a particular disease $d$ and a particular gene $g$.

\paragraph{Entropy.} We compute the entropy $H(g)$ of a gene $g$ in order to measure the specificity of $g$ in terms of the diseases it co-occurs with. If $g$ co-occurs with only a few specific diseases, this results in low entropy and high specificity. Co-occurrence with many diseases yields high entropy and low specificity. We compute the entropy of $g$ as
%\begin{gather}\label{equ:shannonentropy}
$H(g) = -\sum_d^D p(d|g) \cdot \log_2 p(d|g)$, %H(g) = -\sum_d^D p_{d}(d,g) \cdot \log_2(p_{d}(d,g))
% old notation
%p(g|d) = \frac{|T_{dg}| }{\sum_{g'}^{G}|T_{dg'}|} %\qquad p_{d}(d,g) = \frac{|T_{dg}|}{\sum_{d'}^{D}|T_{d'g}|}
%new notation
%p(g|d) = \frac{\#\langle d,\mathit{coocc},g\rangle}{\sum_{g'}^{G}\#\langle d,\mathit{coocc},g'\rangle}
where $p(d|g) = |T_{dg}|/|T_g|$.  %\qquad p_{d}(d,g) = \frac{|T_{dg}|}{\sum_{d'}^{D}|T_{d'g}|}
%\end{gather}
Analogously, we compute the entropy/specificity $H(d)$ of a disease $d$ in terms of the genes it co-occurs with.%\hen{During feature analysis, i mentioned that the entropy of a disease is a very bad feature, since its value is equal for each gene. (logically) Maybe we can take out this feature here. }

\paragraph{Co-occurrence Frequencies.}
This feature group combines relative co-occurrence frequencies of a disease-gene pair $(d,g)$: 
\begin{equation}
 \label{equ:coocc1}%\\
\textit{Occ}(d,g) = \frac{|T_{dg}|}{\underset{d' \in D}\max{\ |T_{d'}|}}
%%\textit{Occ}_{d}(d,g) = \frac{|T_{dg}|}{|T_{d}|-|T_{dg}|} \qquad
%%\textit{Occ}_{g}(d,g) = \frac{|T_{dg}|}{|T_{g}|-|T_{dg}|} \label{equ:coocc23}
\end{equation}
Besides the normalization given in Equation \eqref{equ:coocc1}, two other alternatives are used. In all variants, $\textnormal{Occ}(d,g)$ measures the strength of the connection of $d$ and $g$. Ranging from 0 to 1, small values indicate a weak connection, whereas larger values indicate a strong connection. %Next to the normalization in Equation \eqref{equ:coocc1}, we capture the individual degree of specificity of $d$ with respect to $g$, and vice versa using different normalizations.

\paragraph{Grades.}
This feature group consists of two features which capture a normalized frequency of triples that contain $d$ or $g$, respectively:  
%Each feature value is normalized by its corresponding maximum: 
$\textit{Grade}(d) = |T_{d}| / \underset{d' \in D}\max{\ |T_{d'}|}$ and $\textit{Grade}(g) = |T_{g}| / \underset{g' \in G}\max{\ |T_{g'}|}$.
%
%\begin{equation}
%\textit{Grade}(d) = \frac{|T_{d}|}{\underset{d' \in D}\max{\ |T_{d'}|}} \qquad \textit{Grade}(g) = \frac{|T_{g}|}{\underset{g' \in G}\max{\ |T_{g'}|}}
%\end{equation}
%
\paragraph{Odds Ratio} is used to assess the degree of association between $d$ and $g$:
\begin{equation}\label{equ:odds}
\textit{Odds}(d,g) = \frac{|T_{dg}|\cdot(|T|-|T_{dg}|)}{(|T_{dg}|-|T_{d}|)\cdot(|T_{dg}|-|T_{g}|)}
\end{equation}
The higher $\mathit{Odds}(d,g)$, the stronger the association between $d$ and $g$. $\mathit{Odds}(d,g)=0$ can only be achieved if $|T_{dg}|=0$.

\paragraph{TF-IDF.} In order to assess the relevance of a gene $g$ for a disease $d$, we apply the $\mathit{tfidf}$ metric from information retrieval which takes term frequency ($\mathit{tf}$) and inverted document frequency ($\mathit{idf}$) into account \cite{Manningetal2008}: $\mathit{tfidf}(d,g) = \mathit{tf}(d,g)\cdot\mathit{idf}(g)$. Considering a disease as a ``bag of genes'', $\mathit{tf}(d,g)$ is equivalent to $|T_{dg}|$, while $\mathit{idf}(g)$ can be computed in terms of Equation (\ref{equ:idf}):
\begin{align}\label{equ:idf}
\textit{idf}(g) &= \log\left(\frac{|D|}{\sum_{d \in D} f(d,g)}\right),\ \textnormal{where}\ 
f(d,g) =\left\{\begin{array}{ll}
1 & \textnormal{if}\ |T_{dg}| > 0\\
0 & \textnormal{else}
\end{array}\right.
\end{align}
High values of $\mathit{tfidf}(d,g)$ indicate that $g$ is mentioned frequently in the context of $d$, but still sufficiently specific to be informative for $d$, which we expect to be indicative of a relevant association between $d$ and $g$. 

\subsection{Graph-based Features}\label{sec:genenetworkbased}
In contrast to the previously described feature groups which take a
disease and a gene into account, GBF features are calculated \textit{independently} of a particular disease in that they are entirely based on the gene interaction graph, i.e., the set of triples $I \subset T = \{\langle g,\mathit{interact},g'\rangle|g,g' \in G\}$. %We define $T_{genes}$ as the set of all gene interaction triples in the database.

\paragraph{Path Signatures.}
Each gene $g$ is described in terms of a ``bag of (outgoing) path signatures'', $S_{\textnormal{out}}(g) \subset I = \{\langle g', \mathit{interact}, g''\rangle|g=g'\}$, which have been constructed by joining individual edges in the gene interaction network (cf.\ Section \ref{sec:sysarc}, postprocessing step (3e)). We use \textit{interact} as a placeholder for all predicates constructed in this process. The strength of interaction between a pair $\langle g, g'\rangle \in S_{\textnormal{out}}(g)$ is weighted by the $\mathit{tfidf}$ metric, expressing $\mathit{tf}(g,g')$ and $\mathit{idf}(g')$ analogously to the definitions in Section \ref{sec:cooccurrencebased} (cf.\ Equation(\ref{equ:idf})).
Path signatures encoding an important relation between $g$ and $g'$ in terms of a high $\mathit{tfidf}$ value are considered useful for predicting novel disease-gene associations in cases where no direct evidence from co-occurrence relations is available yet.
%
\begin{comment}
\begin{gather}\label{equ:sigtfidf}
\textit{tfidf}_{\textnormal{sig}}(s,g,G) = \textit{tf}_{\textnormal{sig}}(s,g) \cdot \textit{idf}_{\textnormal{sig}}(s,G)\\[2ex]
%
\textit{tf}_{\textnormal{sig}}(s,g) = \frac{\sum\limits_{s' \in S(g)} f(s,s')}{\max\left\{\sum\limits_{s' \in S(g)} f(s,s') : s\in S(g)\right\}}, \textnormal{where}\
f(s,s') = \left\{\begin{array}{ll}
1 & \textnormal{if}\ s=s'\\ 
0 & \textnormal{else}
\end{array}\right.\\
%
\textit{idf}_{\textnormal{sig}}(s,G) = \log\left(\frac{|G|}{\sum\limits_{g \in G} f(g,s)}\right), \textnormal{where}\
f(g,s) =\left\{\begin{array}{ll}
1 & \textnormal{if}\ s \in S(g) \\
0 & \textnormal{else}
\end{array}\right.
\end{gather}\rom{I propose to formulate the indicator functions with
  $\mathbf{1}_{s=s'}$ and analogously for $S(g)$.}\rom{It is really not clear to me what an outgoing signature
  could be and how set relations are defined on them. I assume these
  should be sets of edges (or nodes?), but I am not sure.}
%
%Since the number of possible features grows exponentially with the number of genes and the path length, we set the maximum path length to two. 
\end{comment}

\paragraph{Gene Connectivity.}\label{sec:genepresence}
This feature group describes the connectivity of a gene within the graph. Analogously to $S_{\textnormal{out}}(g)$ above, we define $S_{\textnormal{in}}(g,l)$ and $S_{\textnormal{out}}(g,l)$ as lists of all incoming and outgoing signatures of path length $l$, respectively. Further, $L$ denotes the maximum path length. %For all experiments reported in this paper, we set $L=2$. %\marginpar{maybe mention earlier}
Based on these definitions, we count the number of incoming and outgoing signatures for each path length $1 \leq l \leq L$ separately, as given in (\ref{equ:presenceout}), and by accumulation over all path lengths. In these features, higher values indicate a higher connectivity of $g$ in the network.
\begin{gather}\label{equ:presenceout}
\textit{Out}_l(g) = \frac{|S_{\textnormal{out}}(g,l)|}{\underset{g' \in G}\max{\ |S_{\textnormal{out}}(g',l)|}} \quad
\textit{In}_l(g) = \frac{|S_{\textnormal{in}}(g,l)|}{\underset{g' \in G}\max{\ |S_{\textnormal{in}}(g',l)|}}
\end{gather}
%
%\begin{gather}
%\textit{Presence}(g) = \frac{\textit{Out}(g)+\textit{In}(g)}{\max\{(\textit{Out}(g')+\textit{In}(g'')) : g',g'' \in G\}}\\
%\textit{Out}(g) =\sum_{l=1}^{L} |S_{\textnormal{out}}(g,l)| \qquad \textit{In}(g) = \sum_{l=1}^{L} |S_{\textnormal{in}}(g,l)| \nonumber
%\end{gather}
%
We also measure the ratio of outgoing and incoming signatures per gene in terms of $\mathit{IORatio}(g)=|S_{out}(g)|/|S_{in}(g)|$. 
%
%\begin{gather}\label{equ:presenceratio}
%\textit{IORatio}(g) = \frac{\max(1,\textit{Out}(g))}{\max(1,\textit{In}(g))}
%\end{gather}
%
If $\mathit{IORatio}(g)>1$, $g$ has a manipulating
role in the network; otherwise, $g$ tends to be manipulated by other
genes.

\section{Experimental Evaluation}

\subsection{Experimental Settings}

\paragraph{Gold Standard.}
The \emph{Genetic Testing Registry} (GTR; \cite{rubinstein2012nih}) is a manually curated database for results from biomedical experiments, mostly at the intersection of Mendelian disorders and human genes. Our GTR dump contains 5,800 disease-gene associations built from 4,200 diseases and 2,800 genes.

\paragraph{Training and Testing Data} are created from GTR as follows: All disease-gene associations in GTR are considered as positive examples. For each disease in GTR, we additionally generate the same amount of negative training examples by pairing the disease with genes that co-occur in Medline but are not attested in GTR as a valid disease-gene association. The resulting data set is split into 80\% used for training and 20\% for testing. The training set contains 3,665 diseases with 1,781 negative and 1,884 positive examples (i.e., associated genes). The test set comprises 910 diseases with 440 positive and 470 negative examples.

\paragraph{Experimental Procedure.} We train an SVM classifier using
an RBF kernel \cite{19cortes95support} and apply grid search for meta-parameter optimization based on the LibSVM\footnote{\url{http://www.csie.ntu.edu.tw/~cjlin/libsvm}} and WEKA\footnote{\url{http://www.cs.waikato.ac.nz/ml/weka/}} toolkits. The trained
model is applied to the task of predicting genes that are associated
with a given disease. We evaluate the model on the GTR test set
described above, reporting precision, recall and $\text{F}_1$ score.

\subsection{Results and Discussion}

\begin{table}
\centering
\small
\begin{minipage}[b]{0.45\textwidth}
\begin{tabular}{lccc}
\toprule
\textbf{Feature Group} & \textbf{Prec.} & \textbf{Rec.} & \textbf{$\text{F}_1$} \\
\midrule
Entropy & 63.2 & 72.5 & 67.5 \\
Co-occurrence & 82.4 & 69.0 & 75.1 \\
Grade & 62.8 & 82.6 & 71.4 \\
Odds Ratio & 77.9 & 43.8 & 56.1 \\
TF-IDF & 85.3 & 62.8 & 72.3 \\\midrule
Path Signatures & 66.6 & 75.6 & 70.8 \\
Connectivity & 62.3 & 78.3 & 69.4 \\
\bottomrule
\end{tabular}
\end{minipage}%
\begin{minipage}[b]{0.45\textwidth}
\begin{tabular}{lccc}
\toprule
\textbf{Feat.\ Combination} & \textbf{Prec.} & \textbf{Rec.} & \textbf{$\text{F}_1$} \\
\midrule
CBF & \textbf{89.6} & 79.8 & 84.4 \\
CBF+Connectivity & 89.1 & 82.2 & 85.5 \\
CBF+Conn+Best50Sig & 89.4 & \textbf{84.9} & \textbf{87.1} \\
CBF+Conn+Best100Sig & 88.1 & 83.3 & 85.7 \\
CBF+Conn+Best200Sig & 88.2 & 84.1 & 86.1 \\
\midrule
Baseline & 87.7 & 41.5 & 56.3 \\
\bottomrule
\end{tabular}
\end{minipage}
\caption{Evaluation results of classification models based on
  indiviudal feature groups (left) and feature group combinations
  (right) on GTR testing data}
\label{tab:exp-results}
\end{table}
Evaluation results are reported in Table \ref{tab:exp-results}. The
left part displays classification performance of individudal feature
groups; in the right part, testing performance of feature combinations
(as selected by cross-validation on the training data) are shown. The
baseline refers to the performance of a single-feature classifier
relying on co-occurrence counts as described in Equation \eqref{equ:coocc1}.
%\rom{So, in the actual methods, the features are boolean, but in the baseline, there is only one  feature which is numerical? Why no baseline of TEES relations of path-length one? (in-document)}

The results clearly indicate a positive impact of both cooccurrence-based and graph-based features, as all feature combinations yield an increase over the baseline in both precision and recall. The CBF combination achieves highest overall precision, whereas connectivity and signature features improve recall (at the expense of slight losses in precision). As for path signatures, it is most effective to select only a small number of individual paths. We determine the best 50 path signature features based on information gain \cite{kullback1951information}. In the best-performing configuration (CBF+Conn+Best50Sig), our system outperforms the co-occurrence baseline by 1.7 points in precision and 43.4 points in recall.% and close to 30 points in F-Measure. 

Increasing the recall relative to the co-occurrence baseline is a key prerequisite towards our goal of discovering novel disease-gene associations. Given that the GTR gold standard is relatively small and slightly biased towards Mendelian disorders, we subject the best-performing model to another evaluation in a practical use case, as described in the next section.

\subsection{Case Study: Pulmonary Fibrosis}

\begin{table}
\small
\centering
\begin{tabular}{cccccc}
\toprule
 & Correct & Plausible Candidate & Incorrect \\\midrule
Count (Percentage) & 77 (38.5\%) & 79 (39.5\%) & 44 (22.0\%)\\ 
\bottomrule
\end{tabular}
\caption{Preliminary results of manual evaluation of 200 gene candidates predicted for pulmonary fibrosis.}
\label{tab:casestudyres}
\end{table}

In this experiment, our classification model was applied to the entire Medline corpus in order to predict genes related to \textit{pulmonary fibrosis} (PF). The resulting hits were sorted by their corpus frequency and the 200 most frequent candidates were manually evaluated by a biomedical expert (who is not a PF researcher, though). Table \ref{tab:casestudyres} shows the preliminary results of this analysis: 38.5\% of the predictions are unanimously correct, whereas only 22\% are clear errors. The missing mass is due to candidates for which hints were found that the gene may be associated with PF through relevant mechanisms or pathways. Thus, these candidates constitute plausible hypotheses which need further investigation by a PF expert.
Main sources of erroneous predictions are false co-occurrences (e.g., due to negation contexts) or false positives as produced by the gene recognition component. Some errors of the latter type may be eliminated by incorporating the filtering approach proposed by \cite{Hartungetal2014}. In sum, this analysis clearly shows that our system is capable of generating promising candidates worth further investigation.% by biomedical experts.

\section{Related Work}

Three types of approaches have been proposed to tackle the problem of extracting explicitly mentioned disease-associated genes (DAGs) but also generating novel hypotheses from scientific publications.
First, several authors extract DAGs from existing biomedical databases such as GeneSeeker \cite{van2003new} or PolySearch \cite{cheng2008polysearch}. Pi{\~n}ero et al.\ developed DisGeNET \cite{pinero2015disgenet}, a database quantifying the degree of disease-gene associations by a combination of different sources of evidence, with textual co-occurrence being one of the main sources. Obviously, these approaches lack the ability to discover new target hypotheses. Second, text mining techniques have been considered as an alternative and are mostly based on textual co-occurrence (sentence or document-based). Such systems can be optimized on precision \cite{chun2006extraction} or recall \cite{pletscher2014diseases}. Al-Mubaid presents a technique using various information-theoretic concepts to support the co-occurrence-based extraction \cite{al2005new}.
Third, a promising alternative to overcome mere textual co-occurrence is to aggregate knowledge across single publications (cf.\ \cite{van2013large}) into larger interaction graphs, as we also do in our approach. Nevertheless, the knowledge extraction to build those interaction graphs often relies on text mining techniques and natural language processing methods as in the BITOLA system \cite{baud2003improving} or in the approaches of Wren et al.\ \cite{wren2004knowledge} and Wilkinson et al.\ \cite{wilkinson2004method}. Closely related to our approach is the one by {\"O}zg{\"u}r et al.\ \cite{ozgur2008identifying} who extract interaction paths from dependency networks and rely on graph centrality measures to rank proteins for a given disease. Contrary to our model, they do not use complex features extracted from the graph and do not combine different types of features.

\section{Conclusions and Outlook}

In this paper, we have presented a system and a model for predicting disease-gene associations from biomedical text, using a combination of features based on disease-gene co-occurrences and gene interactions that are represented in a graph database. In a classification experiment against a manually curated database used as gold standard, we were able to demonstrate the effectiveness of both types of features, outperforming a plain co-occurrence baseline by more than 30 points in $\text{F}_1$ score. Moreover, preliminary investigation of a practical use case from pharmaceutical industry suggests that almost 80\% of the candidates predicted by our model are plausible and may support pharmaceutical researchers in hypothesis generation. In future work, we will carry out a more detailed evaluation of the case study and supplement our classification approach by a ranking model that not only separates positive and negative candidates but also reflects relative differences in these candidates' plausibility.

\bibliographystyle{splncs03} % Style BST file
\bibliography{bibliography}      % Bibliography file (usually '*.bib' )

\end{document}